# SenCos-GEM: SENet-Calibrated and Law-of-Cosines-Constrained Geometry-Enhanced Molecular Representation for Property Prediction

Tianming Han[1,2,#], Li Zhang[3,#], Qi Zhao[1,*]

[1]School of Computer Science and Software Engineering, University of Science and Technology Liaoning, Anshan, 114051, China

[2]EFREI Engineering School of Digital Technologies, Paris-Panthéon-Assas University, Villejuif, 94800, France

[3]School of Life Science, Liaoning University, Shenyang, 110036, China

[#]These authors contributed equally to the paper as first authors.

*Corresponding author

**Email:** zhaoqi@lnu.edu.cn

**Abstract:**

Effective molecular representation learning is crucial for accurate molecular property prediction. Recently, numerous self-supervised learning (SSL) approaches leveraging three-dimensional (3D) graph neural networks (GNNs) have been developed to capture comprehensive 3D structural information of molecules, which is essential for drug discovery. However, existing methods lack explicit physical constraints and are highly susceptible to geometric noise induced by coarse empirical force fields during large-scale pre-training. Furthermore, they overlook dynamic feature modulation during downstream adaptation, often resulting in catastrophic forgetting and negative transfer. To address these limitations, we introduce SenCos-GEM, a novel explicitly decoupled geometry-enhanced molecular representation learning framework that incorporates SENet-calibrated and law-of-cosines-constrained enhancements. SenCos-GEM employs a physics-guided geometric consistency loss based on the law of cosines to derive high-fidelity and mathematically invariant 3D spatial priors. In addition, lightweight Squeeze-and-Excitation (SE) modules are integrated into the backbone as task-specific adapters. These are complemented by an advanced dual-modulation prediction head that integrates Feature-wise Linear Modulation (FiLM) and SENet mechanisms to enable dynamic feature recalibration. SenCos-GEM demonstrates highly competitive performance across diverse classification and regression tasks on MoleculeNet benchmark, establishing new state-of-the-art results specifically on 3D conformation-sensitive regression tasks, such as FreeSolv, Lipophilicity, and QM9, achieving relative error reductions of 12.9% (RMSE), 5.3% (RMSE), and 8.2% (MAE), respectively. Moreover, our model exhibits superior capability in distinguishing stereoisomers and discriminating conformational perturbations, underscoring its robust spatial modeling performance. Collectively, SenCos-GEM represents a significant breakthrough in accurate molecular property prediction.

## 1. Introduction

The determination of molecular properties is of paramount importance in drug discovery, as these properties are closely linked to the biological activity of drugs and their pharmacokinetic profiles, including absorption, distribution, metabolism, and excretion (ADME) [1]. In recent years, machine learning, particularly deep learning approaches, has achieved remarkable success in the field of drug molecular property prediction [2–5]. The corresponding algorithms have also been extended to other molecular design tasks, such as chemical reaction prediction, molecular generation, and conformation generation [6–8]. Furthermore, advanced machine learning techniques have demonstrated substantial potential in anomaly detection and fault diagnosis within physicochemical industrial processes. Notably, for downstream tasks in molecular design, the primary objective is to derive effective molecular representations. Comprehensive representations serve as a core factor in mitigating modeling biases across diverse molecular discovery tasks, as they directly govern the construction of efficient mappings from the molecular chemical space to the functional space. Researchers have developed various molecular representation methods, such as mapping molecules to fixed-dimensional vectors using specific substructures [9] or directly encoding molecules through sequence models based on Simplified Molecular Input Line Entry System (SMILES) strings [10]. Moreover, since graph structures intuitively capture molecular topologies, numerous graph neural network (GNN)-based models have been developed and have yielded promising results [11–14].

However, traditional graph representations, which treat atoms as nodes and chemical bonds as edges, are confined to capturing the two-dimensional (2D) topological structure of molecules and fail to encode the three-dimensional (3D) geometric information that is more critical to molecular properties [15]. With the deepening understanding of structure-property relationships, the integration of 3D geometric information into GNNs has become a consensus [16–19]. Nevertheless, current 3D representation models still face two major bottlenecks. First, at the spatial encoding level, existing 3D GNNs typically aggregate spatial features in a static or isotropic manner, disregarding specific geometric relationships such as the complex modulation effects of continuous bond lengths and bond angles on node states. A representation model capable of dynamically modulating spatial interactions according to local geometric properties would substantially enhance its capacity to perceive authentic molecular conformations. Second, even when these models successfully extract comprehensive structural representations, a critical bottleneck arises in the downstream property prediction stage. Most existing models rigidly map high-dimensional graph representations to property endpoints via a static multi-layer perceptron (MLP). This rigid readout mechanism lacks dynamic feature modulation capabilities, rendering it insufficiently adaptable when transforming general geometric features into highly task-specific biophysical and chemical properties. Consequently, constructing a prediction head based on global graph features that can adaptively scale and recalibrate channel-wise responses would markedly improve the model's predictive performance [20].

In addition to the rigidity inherent in downstream prediction architectures, the scarcity of experimental data constitutes another major limitation. To address the challenge of limited labeled data, self-supervised learning (SSL) pre-training on large-scale unlabeled datasets has emerged as the standard paradigm for acquiring general chemical knowledge [2]. However, prevailing pre-training and fine-tuning frameworks commonly encounter two significant dilemmas. First, during the pre-training phase, merely reconstructing masked geometric targets frequently fails to explicitly constrain the directional properties of latent hyperspace, thereby rendering the extracted representations highly sensitive to microscopic conformational perturbations. Second, when adapting fully pre-trained networks to specific downstream tasks with limited samples, negative transfer or catastrophic forgetting frequently occurs owing to the overly rigid pre-trained weights. Existing models treat all extracted latent feature channels uniformly, lacking a flexible, task-oriented adaptive mechanism. As a result, they are unable to effectively recalibrate feature responses for different target properties during fine-tuning without compromising the pre-trained structural priors.

In this study, we propose a robust and explicitly decoupled molecular representation learning framework, termed SenCos-GEM. This framework separates general geometric pre-training from highly adaptive downstream task modulation. First, in the large-scale self-supervised pre-training stage based on GeoGNN backbone, we introduce a novel objective function that integrates a geometric consistency loss based on cosine similarity. As conceptually illustrated in Figure 1, this physical constraint explicitly regularizes the latent space, prompting the model to align representations directionally and extract highly robust and invariant 3D molecular structure representations from diverse conformations. Second, to overcome the catastrophic forgetting problem during downstream adaptation, we innovatively embed SE modules into each encoder layer. Critically, these intra-layer SE modules are intentionally skipped during the pre-training stage and are only introduced with random initialization during the fine-tuning stage. As lightweight task-specific adapters, they can adaptively recalibrate channel-level feature dependencies for specific property endpoints while preserving the rich pre-trained geometric priors. Finally, we thoroughly reconstruct the downstream decoding mechanism by developing a novel SenCos-GEM prediction head. This customized prediction head overcomes the limitations of traditional static MLP by integrating Feature-wise Linear Modulation (FiLM) with an additional Squeeze-and-Excitation (SE) network, which dynamically scales, shifts, and explicitly screens task-related feature channels prior to the final prediction. Extensive experiments on the standard MoleculeNet benchmark demonstrate that our framework delivers highly competitive predictive performance across diverse classification and regression tasks, establishing new state-of-the-art (SOTA) results specifically on 3D conformation-sensitive properties and large-scale screening challenges. The main contributions of this paper are as follows:

(1) In the self-supervised pre-training stage, a geometric consistency loss based on cosine similarity is designed, enabling our model to achieve geometric alignment of representations in the hyperspace and learn highly robust and invariant 3D molecular structure representations.

(2) An innovative task-specific adaptation strategy is proposed that employs intra-layer SE modules, which do not participate in pre-training and are randomly initialized as lightweight adapters, to recalibrate feature channels for specific downstream targets, thereby effectively alleviating catastrophic forgetting and overfitting on small-scale datasets.

(3) A new SenCos-GEM dual-modulation downstream prediction head is developed. By finely fusing FiLM [21] and SENet [22] mechanisms at the prediction level, our model overcomes the rigidity of static MLPs, dynamically modulates message passing according to complex geometric contexts, and optimizes global graph representations, thereby significantly enhancing property-specific prediction accuracy.

## 2. Related work

Molecular representation learning is fundamental to AI-driven drug discovery. Early GNN models, such as D-MPNN [23] and AttentiveFP [24], primarily focus on extracting 2D topological features, inherently overlooking critical 3D spatial dynamics. To capture molecular conformations, geometry-enhanced models, including SGCN [25], DimeNet [26], and GEM [27], have been developed to incorporate pairwise atomic distances and bond angles. Concurrently, to mitigate the scarcity of labeled data, multi-modal and multi-view SSL frameworks, such as GROVER [28], GraphMVP [29], and 3D Infomax [30], have been proposed to align and integrate 2D topological structures with 3D geometric information. Despite their robust representational capabilities, these large-scale 3D pre-training approaches typically rely on computationally generated conformations such as MMFF94, which often introduce significant geometric noise. Lacking explicit mathematical constraints to filter such noise, most existing frameworks are susceptible to memorizing spatial artifacts, frequently resulting in negative transfer during fine-tuning on highly conformation-sensitive tasks. Another key research area involves the aggregation and fusion of graph features for downstream property prediction. In most molecular representation pipelines, a static MLP is commonly employed to decode global graph embeddings. However, recent studies have highlighted that static readouts often fail to capture the complex, task-specific inductive biases required for diverse chemical endpoints. To enable more adaptive feature fusion, dynamic feature modulation techniques have garnered increasing attention. For example, SENet introduces a channel-wise recalibration mechanism, while FiLM facilitates conditional affine transformations. Recent advances in parameter-efficient fine-tuning, such as AdapterGNN [31] and optimal tuning strategies, along with dual-channel attention networks like DFusMol [32], underscore the value of task-specific feature recalibration and fusion. Nevertheless, the seamless integration of these advanced dynamic gating and conditioning mechanisms into 3D molecular representation learning frameworks remains largely underexplored. Most current 3D models continue to treat extracted spatial feature channels uniformly during downstream adaptation, limiting their ability to dynamically decouple general geometric priors from task-specific macroscopic property predictions.

## 3. Model framework

This section details the architecture and theoretical foundations of our proposed SenCos-GEM framework for geometry-enhanced molecular representation learning. Given that 3D spatial conformations of molecules play a pivotal role in their macroscopic physical, chemical, and biological properties, our framework has been specifically designed to capture intrinsic geometric and topological information. This framework follows a two-stage paradigm: a physics-guided self-supervised pre-training phase to learn the foundational spatial manifold, followed by a downstream adaptive fine-tuning phase optimized for specific molecular property predictions. Overall, the framework of SenCos-GEM comprises three key components, including SE-GeoGNN backbone, physics-constrained geometric pre-training tasks, and an advanced task-specific dual-modulation prediction head.

### 3.1 SE-GeoGNN

Standard graph neural networks primarily operate on 2D topological structures, treating atoms as nodes and chemical bonds as edges. This paradigm is inherently incapable of capturing the rich three-dimensional spatial dynamics induced by bond angles and torsions. To endow the message-passing mechanism with explicit sensitivity to three-dimensional continuous space, we adopt the geometry-based GNN architecture (GeoGNN) [27] as our foundational architecture. Building upon this, our proposed SE-GeoGNN simultaneously models the interaction effects among atoms, chemical bonds, and bond angles. Specifically, as depicted in Figure 2(a), the architecture operates in parallel on two distinct yet mathematically interconnected computational graphs, which are the atom–bond graph $G$ and the bond–angle graph $H$.

To establish a robust and universal geometric foundation, we adopt a phased feature modulation strategy [33]. In the self-supervised pre-training phase, the backbone of GeoGNN employs a simplified foundational message-passing pathway by intentionally skipping the layer-wise channel recalibration mechanism. This deliberate design enables the original layers to directly extract universal intrinsic spatial regularities from three-dimensional coordinates, ensuring that the foundational embeddings strictly adhere to Euclidean topology. In the downstream fine-tuning phase, we explicitly activate SENet mechanism in each GeoGNN block and introduce it via random initialization (wherein the weights and biases for final linear layer of the gating network are initialized to zero, rendering it an identity mapping in the initial training stages). This approach introduces dynamic channel adaptability, enabling the network to adaptively recalibrate feature responses according to specific downstream tasks and handle complex task-specific data distributions while fully preserving the geometric priors obtained from pre-training [34].

Formally, the dual-graph message passing and the layer-SE mechanism are mathematically defined as follows. GeoGNN models the complex molecular spatial environment through two interconnected computational graphs. Let $\mathcal{V}$ be the set of atoms, $\mathcal{E}$ be the set of chemical bonds, and $\mathcal{A}$ be the set of bond angles. The macroscopic atom-chemical bond graph is defined as $G = (\mathcal{V}, \mathcal{E})$, where $u \in \mathcal{V}$ is a node and $(u, v) \in \mathcal{E}$ is an edge. The microscopic chemical bond-bond angle graph is defined as $H = (\mathcal{E}, \mathcal{A})$, where $(u, v) \in \mathcal{E}$ is a node and the bond angle $(u, v, w) \in \mathcal{A}$ connecting two adjacent bonds is an edge.

The hidden representations of atom $u$ and bond $(u, v)$ in the $k$-th message passing iteration are $h_u^{(k)}$ and $h_{uv}^{(k)}$, respectively, which are initialized from the initial spatial features $x_u, x_{uv}, x_{uvw}$. The iterative update formula for bond $(u, v)$ in the graph $H$ is:

$$a_{uv}^{(k)} = \mathrm{AGGREGATE}_H^{(k)}\Big( \big\{ (h_{uv}^{(k-1)}, h_{uw}^{(k-1)}, x_{uvw}) : w \in \mathcal{N}(u) \big\} \cup \big\{ (h_{uv}^{(k-1)}, h_{vw}^{(k-1)}, x_{uvw}) : w \in \mathcal{N}(v) \big\} \Big) \tag{1}$$

$$h_{uv}^{(k)} = \mathrm{COMBINE}_H^{(k)}\Big( h_{uv}^{(k-1)}, a_{uv}^{(k)} \Big) \tag{2}$$

Subsequently, the intermediate spatial representation vector $\tilde{h}_u^{(k)}$ of atom $u$ in the graph $G$ aggregates the updated geometric bond information:

$$a_u^{(k)} = \mathrm{AGGREGATE}_G^{(k)}\Big( \big\{ (h_u^{(k-1)}, h_v^{(k-1)}, h_{uv}^{(k)}) : v \in \mathcal{N}(u) \big\} \Big) \tag{3}$$

$$\tilde{h}_u^{(k)} = \mathrm{COMBINE}_G^{(k)}\Big( h_u^{(k-1)}, a_u^{(k)} \Big) \tag{4}$$

After obtaining the intermediate node representations $\tilde{h}_u^{(k)}$, we introduce a layer-wise channel recalibration mechanism in each GeoGNN block to enhance the adaptive responsiveness of node representations to the graph-level context. For the molecular graph $G$, we first perform graph-global average pooling on the intermediate representations of all nodes to obtain the graph-level statistical vector $\tilde{h}_G^{(k)}$ corresponding to that layer:

$$\tilde{h}_G^{(k)} = \frac{1}{|\mathcal{V}_G|}\sum_{u\in\mathcal{V}_G}\tilde{h}_u^{(k)} \tag{5}$$

On this basis, the channel weight vector $s_G^{(k)}$ corresponding to this graph is generated through a two-layer perceptron and a nonlinear gating function:

$$s_G^{(k)} = 2\cdot\sigma(\mathbf{W}_2\mathrm{ReLU}(\mathbf{W}_1\tilde{h}_G^{(k)} + \mathbf{b}_1) + \mathbf{b}_2) \tag{6}$$

where $\sigma(\cdot)$ is sigmoid gating function and $\mathrm{ReLU}(\cdot)$ denotes rectified linear unit activation, while $\mathbf{W}_1$, $\mathbf{W}_2$ and $\mathbf{b}_1$, $\mathbf{b}_2$ are the learnable weight matrices and biases of the two-layer perceptron, respectively. To avoid the newly introduced channel gating from excessively suppressing message-passing results in the early training stage, the weights and biases of the gating network's final linear layer are initialized to zero and the sigmoid output is multiplied by a coefficient of 2. Consequently, at initialization there holds

$$s_G^{(k)} = 2\sigma(0) = 1 \tag{7}$$

thereby making the layer-wise SE module approximately behave as an identity mapping in the early training stage. Subsequently, this graph-level channel weight $s_G^{(k)}$ is broadcast back to all nodes in the graph, and dimension-wise recalibration is performed on the intermediate node representations $\tilde{h}_u^{(k)}$:

$$h_u^{(k,\mathrm{SE})} = \tilde{h}_u^{(k)} \odot s_G^{(k)}, \quad u \in \mathcal{V}_G \tag{8}$$

where $\odot$ denotes element-wise multiplication. Finally, the recalibrated node representation $h_u^{(k,\mathrm{SE})}$ is residually connected with the input representation from the previous layer to obtain the final output of the$k$-th layer:

$$h_u^{(k)} = h_u^{(k,\mathrm{SE})} + h_u^{(k-1)} \tag{9}$$

This design does not learn independent channel gating for each individual node. Instead, it first generates graph-level channel weights based on the overall node responses of same molecular graph and then broadcasts them back to all nodes within the graph. Therefore, this mechanism can adaptively recalibrate each layer's message-passing results according to the graph-level context statistics of current molecular graph while preserving the shared structural priors among nodes within the graph. It should be noted that this layer-wise SE mechanism is embedded inside GeoGNN encoder, with its action positioned after the message-passing update, normalization, activation, and dropout, but before the residual connection.

### 3.2 Physics-guided synergistic self-supervised pre-training

To overcome the spatial noise bottleneck introduced by empirical force fields and equip our model with a profound understanding of 3D manifolds, we propose a physics-guided synergistic self-supervised pre-training paradigm that transcends traditional isolated multi-scale predictions. Specifically, building upon the reconstruction of local and global geometric targets, we introduce an innovative physical constraint mechanism based on the law of cosines. As illustrated in Figure 2(b), this mechanism seamlessly integrates the following five meticulously designed independent tasks: (1) subgraph context prediction; (2) bond length prediction; (3) bond angle prediction; (4) atomic distance prediction; and (5) functional group prediction. By jointly optimizing microscopic local geometric parameters and macroscopic global topology within a unified Euclidean manifold, this physics-driven synergistic mechanism encourages the originally isolated predictive objectives to mutually constrain and self-correct in the latent space. This approach fundamentally prevents passive memorization of coarse conformational coordinates, thereby enabling the network to filter out the intrinsic noise from empirical force fields and distill highly robust, high-fidelity, and mathematically invariant 3D structural priors [35].

During the pre-training stage, we employ a dual-view encoding strategy. For each molecule, our model processes both the original graph view and the context-masked view simultaneously. In the context-masked view, 15% of the atoms and their one-hop geometric context (adjacent bonds and bond angles) are randomly masked. Except for the geometric consistency constraint, the objective functions of all prediction tasks are computed and aggregated across both views to maximize the consistency of learned representations [36].

### 3.2.1 Local spatial structure

Bond lengths and bond angles are fundamental geometric parameters that characterize the rigidity of local molecular structures. To capture these local spatial features, our network employs hidden states to predict continuous bond lengths and normalized bond angles over the complete set of edges and angle triplets. Additionally, to enhance the model's robustness to geometric noise and structural outliers, we employ $\mathrm{Smooth}_{L_1}$ loss rather than conventional mean squared error for these local prediction tasks.

### 3.2.2 Global spatial structure

In addition to local spatial structures, we introduce an atomic distance prediction task to capture the global geometry of molecules. For each molecule, pairwise atomic distance matrices are constructed from three-dimensional coordinates. Given the substantial variance in long-range spatial distances across different flexible conformations, this task is formulated as a robust multi-class classification problem rather than direct regression. Specifically, the continuous Euclidean distances are truncated at 20 Å and uniformly discretized into 31 effective bins of equal width.

### 3.2.3 Chemical semantics and context

In addition to spatial geometry, capturing topological and chemical properties is of critical importance. We introduce a functional group prediction task, in which the prediction of molecular fingerprints and functional group descriptors is formulated as a series of binary classification problems based on global graph representations. Furthermore, we incorporate a subgraph context prediction task that classifies the categorical labels of masked local subgraph contexts into a predefined vocabulary via hash mapping.

### 3.2.4 Geometric consistency based on law of cosines

Independent prediction of local geometric parameters and global distances risks producing mathematically inconsistent 3D manifolds in Euclidean space. Moreover, this unconstrained optimization paradigm substantially exacerbates a critical data quality bottleneck in large-scale 3D pre-training. Generating millions of spatial conformations requires empirical force fields (e.g., MMFF94 [37]), which inevitably yield coarse, low-precision geometric structures relative to rigorous quantum mechanical calculations and contain substantial structural noise [38]. When unconstrained networks blindly regress on these isolated targets, they tend to overfit the specific biases and artifacts of empirical force fields, thereby memorizing coordinate noise rather than learning the intrinsic spatial topology.

To fundamentally address these limitations and protect representation learning process from low-precision bottleneck inherent to empirical force fields, we introduce a novel geometric consistency constraint based on law of cosines [39]. In Euclidean geometry, the lengths of two adjacent bonds and their included angle strictly determine the spatial distance between two endpoint atoms. We design a physical penalty term to minimize the discrepancy between expected squared atomic distance obtained from the distance classification head and the theoretical squared distance derived from predicted bond lengths and bond angles.

Crucially, this physical regularizer elegantly decouples the representation learning process from the absolute precision of training coordinates. Our model no longer passively memorizes the noisy geometric structures generated by MMFF94. Instead, it internalizes the universal and invariant mathematical principles of three-dimensional space. Gradients from this single physical constraint are back-propagated simultaneously to refine the predictions of bond lengths, bond angles, and distances, compelling them to mutually constrain and self-correct. This physically grounded synergy serves as a powerful structural filter, enabling our network to extract highly robust and high-fidelity spatial representations even from inherently coarse geometric structures. To enhance numerical stability and avoid anomalous gradients from square root operations during back-propagation, the constraint is formulated in the squared distance domain and evaluated solely on the original graph view.

### 3.2.5 Static loss aggregation

The overall pre-training objective is formulated as a simple static summation of geometric and semantic losses. In contrast to previous pre-training methods, which often rely on complex heuristic dynamic loss-weighting schedulers to balance independent tasks due to conflicting gradients when fitting noisy spatial data, the physically grounded law-of-cosines constraint introduced herein fundamentally resolves this gradient conflict. Because the previously isolated objective functions now actively interact through the cosine bridge and are subject to mutual mathematical constraints, their gradient optimization directions achieve fundamental alignment within the shared valid geometric manifold. Consequently, this simple static aggregation, when combined with a dynamic warm-up schedule for the law-of-cosines coefficient, naturally yields highly stable and synergistic convergence during pre-training, effectively mitigating local conformational noise without requiring heuristic dynamic weighting mechanisms [40].

### 3.2.6 Mathematical formulation of pre-training objectives

The specific objective functions and the physical constraints for these pre-training tasks are mathematically formulated as follows. Following the definitions of the aforementioned symbols, let $\mathcal{P} = \mathcal{V} \times \mathcal{V}$ be the set of all atom pairs. For each molecule, our model simultaneously encodes both the original graph view and the context-masked view to generate the corresponding representations. Except for the geometric consistency term, all other geometric and semantic tasks are computed separately on the two views and then summed.

For local geometric modeling, our model predicts bond lengths and bond angles over the complete edge set and angle triplet set. To enhance robustness against geometric noise and outliers, the smooth $L_1$ loss is employed rather than mean squared error. Mathematically, for a generic prediction error $x$, the smooth $L_1$ function is defined as a piecewise function:

$$\text{Smooth}_{L_1}(x) = \begin{cases} 0.5x^2, & \text{if } |x| < 1 \\ |x| - 0.5, & \text{otherwise} \end{cases} \tag{10}$$

The bond length regression loss $L_{\text{Blr}}$ is defined as:

$$L_{\text{Blr}} = \frac{1}{|\mathcal{E}|} \sum_{(u,v)\in\mathcal{E}} \left[ \text{Smooth}_{L_1}(\hat{l}_{uv} - l_{uv}) + \text{Smooth}_{L_1}(\tilde{l}_{uv} - l_{uv}) \right] \tag{11}$$

where $\hat{l}_{uv}$ and $\tilde{l}_{uv}$ are the predicted bond lengths under the original graph view and masked view, respectively, and $l_{uv}$ is the ground-truth bond length. For the bond angle task, normalized angle supervision is adopted by scaling the ground-truth angle $\phi_{uvw}$ to $\phi_{uvw}/\pi$. The bond angle regression loss $L_{\text{Bar}}$ is defined as:

$$L_{\text{Bar}} = \frac{1}{|\mathcal{A}|}\sum_{(u,v,w)\in\mathcal{A}}\left[\text{Smooth}_{L_1}(\hat{a}_{uvw} - \phi_{uvw}/\pi) + \text{Smooth}_{L_1}(\tilde{a}_{uvw} - \phi_{uvw}/\pi)\right] \tag{12}$$

where $\hat{a}_{uvw}$ and $\tilde{a}_{uvw}$ are the predicted normalized bond angles under the two views. In the geometric consistency constraint, the physical angle $\hat{\phi}_{uvw}$ is recovered via

$$\hat{\phi}_{uvw} = \pi \hat{a}_{uvw}. \tag{13}$$

To model the global spatial topology, discrete classification of Euclidean distances is performed over all atom pair sets $\mathcal{P}$. The continuous distances are truncated to [0, 20] Å and mapped to 31 classification bins (indices 0, …, 30). Let $d_{uv}$ be the true physical distance, and its discrete label $y_{uv}^{\text{dist}}$ is constructed as:

$$y_{uv}^{\text{dist}} = \left[\frac{\min(d_{uv}, 20)}{20}\cdot 30\right], \quad y_{uv}^{\text{dist}} \in \{0,\ldots,30\} \tag{14}$$

Let $\hat{z}_{uv} \in \mathbb{R}^{31}$ and $\tilde{z}_{uv} \in \mathbb{R}^{31}$ be the logit outputs of the distance head under the original view and the masked view, respectively. The atomic distance classification loss $L_{Adc}$ is defined as:

$$L_{\text{Adc}} = \frac{1}{|\mathcal{P}|}\sum_{(u,v)\in\mathcal{P}}\left[\text{CE}(\hat{z}_{uv}, y_{uv}^{\text{dist}}) + \text{CE}(\tilde{z}_{uv}, y_{uv}^{\text{dist}})\right] \tag{15}$$

where CE denotes the cross-entropy loss.

Let $c_b = \dfrac{20b}{30}$ denote the center of the$b$-th effective bin. The expected distance $\hat{d}_{uv}$ under the original view is obtained via analytical decoding as:

$$\hat{d}_{uv} = \sum_{b=0}^{30} c_b \,\text{Softmax}(\hat{z}_{uv,0:30})_b \tag{16}$$

This expected distance $\hat{d}_{uv}$ is used solely for constructing the subsequent geometric consistency constraint.

In addition to the geometric tasks, our model enhances its perception of chemical semantics through the functional group predictive task. Let $y_{\text{Fg}}$ denote the graph-level supervision target formed by concatenating molecular fingerprints and functional group descriptors, and the functional group prediction loss $L_{\text{Fg}}$ is:

$$L_{\text{Fg}} = \text{BCEWithLogits}(f_{\text{Fg}}(\hat{h}_G), y_{\text{Fg}}) + \text{BCEWithLogits}(f_{\text{Fg}}(\tilde{h}_G), y_{\text{Fg}}) \tag{17}$$

where $\hat{h}_G$ and $\tilde{h}_G$ are the graph-level representations under the original view and masked view, respectively.

When the context-masking task is enabled, our model further introduces the local context prediction loss $L_{\text{Cm}}$. Let $\mathcal{M}$ denote the set of selected target atoms. For each target atom $u \in \mathcal{M}$, a local subgraph context is constructed from its one-hop neighboring atoms and adjacent chemical bonds, and the context category label is obtained via hash mapping $c_u \in \{0, \ldots, C_m - 1\}$. Let $\hat{z}_u^{\text{Cm}}$ and $\tilde{z}_u^{\text{Cm}}$ represent the context classification logits under the original graph view and context-masked view, respectively. Then, the context prediction loss $L_{\mathrm{Cm}}$ is defined as:

$$L_{\mathrm{Cm}} = \frac{1}{|\mathcal{M}|} \sum_{u \in \mathcal{M}} \left[ \mathrm{CE}(\hat{z}_u^{\mathrm{Cm}}, c_u) + \mathrm{CE}(\tilde{z}_u^{\mathrm{Cm}}, c_u) \right] \tag{18}$$

To ensure that the local geometric predictions and global distance predictions satisfy Euclidean geometric relationship, we introduce a geometric consistency constraint based on the law of cosines. Compared with directly adopting the square root form, we construct this constraint in the squared distance domain to enhance numerical stability and avoid abnormal gradients potentially caused by the square root term. Specifically, the geometric consistency loss $L_{\cos}$ is defined as:

$$L_{\cos} = \frac{1}{|\mathcal{A}|} \sum_{(u,v,w) \in \mathcal{A}} (\hat{d}_{uw}^2 - \left( \hat{l}_{uv}^2 + \hat{l}_{vw}^2 - 2\hat{l}_{uv}\hat{l}_{vw} \cos(\pi \hat{a}_{uvw}) \right))^2 \tag{19}$$

It should be emphasized that this term is computed only once based on the predictive results from the original graph view, thereby avoiding redundant calculations on the masked view. The continuous distance $\hat{d}_{uw}$ used in the geometric consistency loss is not obtained via direct regression but is decoded from the discrete probability distribution produced by the distance classification head under the original graph view. Specifically, let $\hat{z}_{uw} \in \mathbb{R}^{31}$ represent the distance classification logits for the atom pair $(u, w)$ under the original graph view, only the outputs corresponding to the first 31 valid categories are extracted, normalized via softmax, and weighted summed with each discrete reference distance value to obtain the continuous expected distance:

$$\hat{d}_{uw} = \sum_{b=0}^{30} c_b \, \mathrm{Softmax}(\hat{z}_{uw,0:30})_b \tag{20}$$

where $c_b = \dfrac{20b}{30}$ , $b = 0,\ldots,30$ denotes the reference distance value corresponding to the $b$-th valid distance category. Thus, $\hat{d}_{uw}$ serve as the continuous distance estimate under the original graph view for constructing the geometric consistency constraint based on the law of cosines.

Summarizing the above tasks, the pre-training objective function $L_{\mathrm{pre}}$ under the default configuration can be expressed as:

$$L_{\mathrm{pre}} = L_{\mathrm{Fg}} + L_{\mathrm{Bar}} + L_{\mathrm{Blr}} + L_{\mathrm{Adc}} + \lambda_t L_{\mathrm{cos}} + L_{\mathrm{Cm}} \tag{21}$$

where $\lambda_t$ is the geometric consistency coefficient, which is gradually warmed up to the preset value during training.

### 3.3 Dual-modulation prediction head

Following the pre-training of geometry-consistent molecular representations, our model undergoes fine-tuning on diverse labeled benchmark datasets to predict specific macroscopic endpoints, such as quantum mechanical energy gaps, solvation energies, or biological toxicities. To effectively transform the general three-dimensional molecular representation vectors into high-precision nonlinear target predictions, we develop an advanced and highly parameterized task-specific prediction head.

To address the complexity of downstream property space, this prediction head incorporates a dual-modulation mechanism integrating FiLM and SENet modules. Specifically, an SE-style channel gating mechanism first recalibrates the graph-level representations by enhancing task-relevant channels and suppressing irrelevant noise. Subsequently, conditioned on the graph-level aggregated edge representations, FiLM layer applies task-specific affine transformations (dynamic scaling and shifting) to the recalibrated global features. The modulated graph-level representations are then fed into an MLP to yield the final target property predictions.

In the downstream fine-tuning stage, unlike many existing models that retain geometric or semantic reconstruction as auxiliary tasks, we design the objective function to optimize solely the target macroscopic properties. This focused design directs our model's full parameter capacity and representational capability toward the primary prediction task, thereby establishing a direct and efficient mapping from the microscopic three-dimensional geometric structures to the macroscopic molecular properties.

Mathematically, the forward propagation of this dual-modulation mechanism is executed via the following operations. In the downstream property prediction stage, the encoder outputs the final node and edge representations after K message-passing layers (where K = 8 in our implementation). Following the notations defined in Section 3.1, we omit the layer superscript K for brevity, and let $\{h_u\}$ and $\{h_{uv}\}$ denote the sets of these final node and edge representations, respectively. Based on the node representations, the graph-level representation $h_G$ is computed as:

$$h_G = \mathrm{LayerNorm}(\mathrm{MeanPool}_{\mathcal{V}}(\{h_u\})) \tag{22}$$

First, an SE-style channel gating mechanism is employed to recalibrate $h_G$ by generating a gating vector $g_G$ to obtain the recalibrated representation $h_G^{\mathrm{SE}}$:

$$g_G = \sigma\left(\mathbf{W}_4 \mathrm{ReLU}\left(\mathbf{W}_3 h_G + \mathbf{b}_3\right) + \mathbf{b}_4\right), \quad h_G^{\mathrm{SE}} = h_G \odot g_G \tag{23}$$

where $\mathbf{W}_3$, $\mathbf{W}_4$ and $\mathbf{b}_3$, $\mathbf{b}_4$ denote the learnable weight matrices and biases of SE gating mechanism, respectively. Subsequently, the edge-level representations of the same molecule are aggregated via mean pooling to obtain the graph-level conditional vector $e_G$:

$$e_G = \mathrm{MeanPool}_{\mathcal{E}}\left(\{h_{uv}\}\right) \tag{24}$$

Based on this graph-level aggregated edge representation, dimension-wise modulation parameters $\gamma(e_G)$ and $\beta(e_G)$ are generated:

$$\gamma(e_G) = \sigma\left(\mathbf{W}_\gamma e_G + \mathbf{b}_\gamma\right), \quad \beta(e_G) = \mathbf{W}_\beta e_G + \mathbf{b}_\beta \tag{25}$$

where $\mathbf{W}_\gamma$, $\mathbf{W}_\beta$ and $\mathbf{b}_\gamma$, $\mathbf{b}_\beta$ represent the corresponding learnable linear projection weights and biases, respectively. Subsequently, FiLM modulation is applied to the recalibrated graph-level representation $h_G^{\mathrm{SE}}$ to obtain the modulated representation $h_G^{\mathrm{mod}}$:

$$h_G^{\mathrm{mod}} = \gamma(e_G) \odot h_G^{\mathrm{SE}} + \beta(e_G) \tag{26}$$

Finally, the modulated graph representation $h_G^{\mathrm{mod}}$ is fed into an MLP to yield the target property prediction value $\hat{y}$:

$$\hat{y} = \mathrm{MLP}\left(h_G^{\mathrm{mod}}\right) \tag{27}$$

This design enables the downstream prediction head not only to highlight the channel dimensions that contribute most to the current task but also to leverage edge-level geometric information for conditional correction of graph-level representation, thereby enhancing the discriminative capability of molecular property prediction.

# 4. Experiments

To comprehensively evaluate the performance of SenCos-GEM, we compare it against several state-of-the-art methods across 15 benchmark datasets from MoleculeNet [41]. In addition, we conduct comprehensive ablation studies to validate the contributions of the core components and design specialized diagnostic experiments to assess our model's ability to distinguish stereoisomers and discriminate conformational perturbations.

## 4.1 Pre-training settings

### 4.1.1 Datasets

For SSL of molecular representations, we pre-train SenCos-GEM on ZINC20 dataset [42], a publicly available database of purchasable drug-like compounds. Specifically, we adopt the standard ZINC20 Drug-like 20M subset, which contains 19489955 unlabeled molecules sourced from Hugging Face repository

(https://huggingface.co/datasets/Pixelatory/ZINC20-Druglike). We randomly select 90% of the samples for training and reserve the remaining 10% for model validation.

#### 4.1.2 Self-supervised learning task settings

In the SSL of 3D geometric structures, we employ RDKit to generate conformations for approximately 19.5 million molecules from ZINC20 database under MMFF94 force field. Geometric features of the molecules, including bond lengths, bond angles, and interatomic distance matrices, are derived from the resulting 3D coordinates.

To enable SenCos-GEM to comprehensively capture local chemical environments and global 3D spatial geometric structures, we adopt a masking strategy to implement multi-scale self-supervised learning tasks. Specifically, we randomly mask 15% of atoms in each molecule along with their associated chemical bonds. Considering the varying feature requirements of different downstream properties, we design targeted pre-training strategies for regression- and classification-oriented tasks, respectively.

For regression-oriented pre-training, the self-supervised learning framework includes five distinct tasks: (1) context masking (Cm), with a vocabulary size of 2400, to capture local atomic environments; (2) functional group prediction (Fg), spanning 494 different functional groups;

(3) bond length prediction (Blr), formulated as a continuous regression task; (4) bond angle prediction (Bar), formulated as a continuous regression task; and (5) interatomic distance prediction (Adc), treated as a multi-class classification task by mapping continuous spatial distances to 30 equal-width bins.

For classification-oriented pre-training, to avoid potential feature interference, we intentionally omit the Cm task and retain only the remaining four tasks (Fg, Blr, Bar, and Adc) to guide the optimization process. Additionally, to mathematically enforce structural validity and preserve the intrinsic topological-geometric relationships among reconstructed bond lengths, bond angles, and interatomic distances, we introduce a geometric consistency constraint based on law of cosines. The weight of this geometric consistency loss is empirically set to 4.5 for regression pre-training and 0.8 for classification pre-training. In both configurations, to ensure stable convergence in the early stages of training, the weights are gradually increased via a linear warmup schedule over 10 epochs.

### 4.2 Molecular property prediction settings

#### 4.2.1 Datasets and splitting strategy

To enable direct comparisons with prior studies, we utilize MoleculeNet benchmark datasets to assess the predictive performance of SenCos-GEM. These encompass nine classification tasks (BACE, BBBP, ClinTox, HIV, MUV, PCBA, SIDER, Tox21, and ToxCast) and six regression tasks involving physicochemical and quantum mechanical properties (ESOL, FreeSolv, Lipophilicity, QM7, QM8, and QM9). Consistent with previous studies, we adopt a scaffold-based splitting strategy for all datasets by partitioning the molecules according to their 2D Bemis-Murcko scaffolds [43]. This approach enforces an 8:1:1 ratio for the training, validation, and test sets, ensuring distinct chemical space distributions between training and test sets and thereby affording a rigorous evaluation of the model's out-of-distribution (OOD) generalization capability on structurally novel compounds [44].

#### 4.2.2 Evaluation metrics

Following the recommendations of MoleculeNet, the average area under the receiver operating characteristic curve (ROC-AUC) is adopted as the primary evaluation metric for binary classification tasks, where higher values indicate superior model performance. For multi-task classification datasets containing unmeasured or missing bioactivity labels, such as Tox21 and PCBA, mask tensors are strictly applied during loss computation and ROC-AUC evaluation to preclude the influence of blank annotations. For regression tasks, root mean square error (RMSE) is used to evaluate ESOL, FreeSolv, and Lipophilicity, whereas mean absolute error (MAE) is employed for the quantum mechanical datasets, namely QM7, QM8, and QM9 [45]. Lower values of these metrics correspond to higher predictive accuracy. The details of these metrics are defined as follows:

$$\text{ROC-AUC} = \int_0^1 TPR(FPR)\,\mathrm{d}FPR \tag{28}$$

$$\text{RMSE} = \sqrt{\frac{1}{n}\sum_{i=1}^{n}(y_i - \hat{y}_i)^2} \tag{29}$$

$$\text{MAE} = \frac{1}{n}\sum_{i=1}^{n}| y_i - \hat{y}_i| \tag{30}$$

#### 4.2.3 GNN architecture

To capture comprehensive spatial representations, the base molecular encoder performs message passing simultaneously on dual geometric graphs (atom–bond graph and bond–angle graph). Adhering to the standard message-passing paradigm, Graph Isomorphism Network (GIN) is employed to define the aggregation and update functions. To further enhance representational capacity and training stability, residual connections, layer normalization, and graph normalization [46] are incorporated into each GNN block. Finally, mean pooling is utilized as the readout function to aggregate node-level hidden states into a global graph-level representation.

In the downstream fine-tuning stage, the layer-wise SE (layer-SE) modules in the encoder generate channel gating based on graph-level pooled statistics with a reduction ratio of 4, and perform channel recalibration on the node representations in each GeoGNN block, thereby sharpening the focus of SenCos-GEM on task-relevant representations. Critically, the excitation scaling factors within SENet module are initialized to 1. This identity-mapping initialization ensures seamless transfer of the pre-trained geometric embeddings at the outset of fine-tuning, while preventing initial gradient shocks from disrupting the acquired chemical knowledge. Additionally, the dropout rate in downstream head is configured typically as 0.2 for regression tasks to mitigate overfitting on continuous targets and 0.0 for classification tasks to preserve sharp class boundaries. Decoupled Adam optimizers [47] are applied separately to pre-trained encoder and newly initialized prediction head to promote stable convergence.

**4.2.4 Baseline models**

We compare SenCos-GEM and its backbone SE-GeoGNN with a variety of SOTA baseline models. The non-pre-trained models include standard GNN architectures such as GIN, GCN, GAT, and GTransformer, as well as classic architectures specifically designed for molecular representation, such as D-MPNN [23] and AttentiveFP [24]. We also compare with geometry-aware models that explicitly encode 3D structural information, including SGCN [25], DimeNet [26], HMGNN [48] and GeoGNN [27]. The pre-trained baselines include D-MPNN, AttentiveFP, N-Gram [49], PretrainGNN [50], GROVER [28], and GEM [27]. Specifically, N-Gram predicts molecular properties by combining node embeddings along short paths in the graph and employs random forest ( $\text{N-Gram}_{\text{RF}}$ ) or extreme gradient boosting tree ( $\text{N-Gram}_{\text{XGB}}$ ) for downstream prediction. PretrainGNN implements a variety of general topology-based self-supervised learning tasks. GROVER combines GNN with Transformer architecture and utilizes two SSL tasks. We report its experimental results under two different network capacities, namely $\text{GROVER}_{\text{base}}$ and $\text{GROVER}_{\text{large}}$ . Finally, GEM serves as the most direct comparative benchmark to evaluate the superiority of SenCos-GEM.

# 5. Results

## 5.1 Overall performance

The overall performance of SenCos-GEM relative to other competitive baseline methods in molecular property prediction is summarized in Tables 1 and 2. To ensure robust and statistically reliable comparisons, all models are evaluated across four independent runs with different random seeds. Results are reported as the mean with standard deviations in parentheses. The experimental results lead to the following key observations.

SenCos-GEM delivers highly competitive performance on the regression benchmarks, frequently attaining state-of-the-art results. These tasks assess macroscopic physicochemical properties and microscopic quantum mechanical properties, which are intrinsically sensitive to three-dimensional geometric structures [51]. Crucially, our framework establishes a novel knowledge-data fusion paradigm that seamlessly integrates explicit physical priors with data-driven 3D geometric embeddings. Compared with the previous strongest baseline GEM, which lacks such explicit cross-dimensional integration, SenCos-GEM substantially reduces predictive errors on particularly challenging tasks. For instance, on FreeSolv dataset, SenCos-GEM lowers the RMSE from 1.877 to 1.634, representing a notable 12.9% relative improvement. This dataset pertains to critical aspects of drug solubility, bioavailability, and pharmacokinetics. Similarly, it reduces the RMSE on Lipophilicity from 0.660 to 0.625, equating to a 5.3% improvement. Admittedly, the default configuration introduces a marginal trade-off on ESOL and QM7. As ESOL is predominantly governed by 2D topologies, aggressive 3D feature modulation risks interference. However, adapting a semantic-focused prediction head improves the RMSE to 0.807 (see Section 6.3). For QM7, the highly limited dataset size restricts the model's capacity to adapt the generalized spatial priors acquired during pre-training to the strict sub-Ångström quantum exactness required by this task. Strikingly, training our intrinsic SE-GeoGNN backbone from scratch, thereby bypassing these pre-trained macroscopic priors, establishes a new state-of-the-art MAE of 56.6 on QM7 (see Section 6.1), decisively surpassing GEM. Lipophilicity serves as a pivotal pharmacokinetic descriptor that governs cell membrane permeability and oral drug absorption in the early stages of drug discovery. Furthermore, on QM9, which is widely regarded as the "ImageNet" of quantum chemistry and a premier benchmark for molecular modeling, the MAE decreases from 0.00746 to 0.00685, yielding an 8.2% relative improvement. Collectively, these findings robustly demonstrate that our framework effectively mitigates geometric noise while achieving profound integration of pre-trained spatial priors with task-specific semantics, thereby enabling precise translation of intricate structural representations into accurate property predictions.

On the classification datasets for biological activity and toxicity, such as BACE, BBBP, Tox21, and PCBA, SenCos-GEM consistently demonstrates robust generalization performance. Admittedly, the pronounced 3D geometric regularization introduces an inherent trade-off, resulting in a marginally lower average ROC-AUC of 0.782 compared with the GEM baseline of 0.791 owing to reduced adaptability on 2D-topology-dominated tasks such as ClinTox, SIDER, and ToxCast. Nevertheless, our model still outperforms classical architectures such as D-MPNN and AttentiveFP while maintaining highly competitive ROC-AUC scores relative to large-scale pre-trained models such as GROVER. Notably, despite this geometric emphasis, our dual-modulation prediction head proves exceptionally effective in complex, large-scale screening scenarios. Our model achieves state-of-the-art performance of 0.874 on PCBA, which is the largest and most challenging high-throughput screening benchmark. This superior result on PCBA robustly validates our core design philosophy, which is explicitly decoupling general geometric pre-training from task-specific modulation effectively mitigates negative transfer on complex tasks, thereby enabling synergistic multi-scale feature integration. By dynamically integrating representations via the advanced downstream prediction head, SenCos-GEM adeptly navigates intricate property landscapes and achieves highly adaptive generalization across diverse chemical domains.

### 5.2 Ability to distinguish stereoisomers

To assess whether SenCos-GEM more effectively encodes molecular chirality, we conduct a stereoisomer discrimination experiment [52]. Molecules with 2, 4, 8, and 16 possible stereoisomers are screened from ZINC20M dataset, followed by ChemDiv-like drug-likeness filtering. For each stereoisomer family, RDKit is employed to enumerate all atom-centered chiral isomers derived from the same achiral molecular scaffold, while non-atom-centered stereoisomers (e.g., E/Z isomers) are excluded.

To ensure a direct correspondence between stereoisomer count and chirality complexity, molecular families with precisely 1, 2, 3, and 4 atom-centered chiral centers are retained, corresponding to a stereoisomer count $N_{\text{isomers}}$ of 2, 4, 8, and 16, respectively. In each group, 500 stereoisomer families are sampled. For every stereoisomer, a three-dimensional conformation is generated and optimized using MMFF94 force field. Molecular representations are subsequently extracted using the official GEM encoder and SenCos-GEM, respectively. To rigorously quantify the capacity to discriminate stereoisomers while strictly eliminating the confounding effect of varying absolute embedding magnitudes across models, all extracted representations are first $L_2$-normalized. Subsequently, pairwise Euclidean distances between $L_2$-normalized representations of different stereoisomers within each family are computed [53]. Larger intra-family normalized distances indicate that the model can generate more directionally distinct representations for molecules sharing identical 2D scaffolds but differing in stereochemical configuration. As illustrated in Figure 3, stereoisomer representations produced by GEM tend to exhibit high spatial similarity, clustering closely with average pairwise normalized distances of 0.007, 0.009, 0.012, and 0.013 within families comprising 2, 4, 8, and 16 stereoisomers, respectively. In contrast, SenCos-GEM markedly enlarges the inter-stereoisomer spatial separation, achieving average normalized distances of 0.089, 0.103, 0.112, and 0.119, which are 12.71, 11.44, 9.33, and 9.15 times those of GEM. Statistical analyses further reveal that the normalized distance distributions for SenCos-GEM are significantly larger than those for GEM across all stereoisomer groups, with $p < 0.001$.

### 5.3 Capability for discriminating conformational perturbations

To further validate our model's ability to capture variations in 3D molecular conformations, we design a diagnostic task in the representation space to distinguish between original and perturbed conformations. Specifically, 500 molecules are selected from ZINC20M, and their 3D conformations optimized by MMFF94 force field serve as the original conformations. Uniform noise is subsequently added to 3D coordinates of each atom to generate the perturbed conformations. Centered noise is employed, with coordinate perturbations sampled uniformly from the interval [−0.25, 0.25] Å. Under this setting, the average RMSD between the original and perturbed conformations is 0.249 Å.

For each molecule, the original and perturbed conformations are independently fed into our model to extract molecular representations, which are then projected onto a 2D space using t-SNE for visualization. Three models are compared: the official GEM; Sen-GEM, an ablated variant without the law-of-cosines constraint; and SenCos-GEM. To quantitatively assess the separation between the original and perturbed conformations in the representation space, Davies–Bouldin index (DBI) [54] is calculated. Mathematically, for a dataset partitioned into $N$ clusters, DBI is defined as:

$$\mathrm{DBI} = \frac{1}{N}\sum_{i=1}^{N}\max_{j \neq i}\left(\frac{S_i + S_j}{M_{i,j}}\right) \tag{31}$$

where $S_i$ denotes the intra-cluster dispersion, computed as the average distance between all samples in cluster $i$ and their respective centroid, and $M_{i,j}$ represents the inter-cluster distance between the centroids of clusters $i$ and $j$. In our diagnostic task involving exactly two clusters, with $N = 2$ representing the original and perturbed sets, this formulation explicitly demonstrates that a lower DBI corresponds to tighter intra-cluster cohesion driven by a smaller numerator $S$, and more pronounced inter-cluster separation reflected by a larger denominator $M$ [55].

As shown in Figure 4, the molecular representations extracted by the official GEM exhibit substantial intermixing between MMFF94-optimized original conformations and their perturbed counterparts, hindering the formation of distinct two-class clusters with a DBI of 4.45. In contrast, both Sen-GEM and SenCos-GEM more effectively distinguish the original and perturbed conformations, demonstrating that enhanced geometric representation learning framework improves the sensitivity to conformational variations. Specifically, Sen-GEM yields a DBI of 1.94, whereas SenCos-GEM further reduces it to 1.60, highlighting that the law of cosines consistency constraint confers superior conformational separability in the embedding space.

These results demonstrate that SenCos-GEM not only encodes local geometric information such as bond lengths and bond angles but also enhances consistency among different geometric prediction targets through the law of cosines constraint, thereby yielding a more structured molecular representation space. Relative to the official GEM, SenCos-GEM exhibits stronger discriminative capability for conformational perturbations. Moreover, compared with Sen-GEM, it attains a lower DBI in the high-dimensional representation space, indicating that this constraint improves overall capacity to model variations in three-dimensional molecular geometries. Crucially, because arbitrary uniform noise inherently breaks valid Euclidean topologies, successfully isolating these perturbed pseudo-conformations proves that SenCos-GEM does not blindly memorize coordinate noise. Instead, it demonstrates that our model has genuinely internalized the rigid boundaries of valid molecular geometry, allowing it to effectively filter and reject unphysical structural anomalies.

# 6. Ablation study

## 6.1 Contribution of SE-GeoGNN

To isolate the inherent representational capability of the proposed SE-GeoGNN backbone from the benefits of self-supervised pre-training, we compare it with various GNN architectures trained from scratch on regression tasks. As shown in Table 3, SE-GeoGNN significantly outperforms the other GNN architectures on most regression tasks. Specifically, compared with the standard GeoGNN backbone, incorporation of the intra-layer SE mechanism reduces RMSE on FreeSolv from 1.857 to 1.765, representing a 5.0% error reduction. Furthermore, it decreases MAE on QM7 from 59.0 to 56.6 and MAE on QM9 from 0.00746 to 0.00695, yielding relative improvements of 4.1% and 6.8%, respectively. These results demonstrate that integrating channel recalibration mechanism into message-passing process substantially enhances the backbone network's ability to dynamically capture complex three-dimensional atom-bond-angle interactions. By adaptively recalibrating node-level responses based on graph-level context, our network extracts more discriminative spatial representations even in the absence of large-scale pre-training.

### 6.2 Impact of cosine constraint pre-training strategy

To evaluate the effectiveness of proposed physics-guided self-supervised learning tasks, we examine various pre-training strategies applied to SE-GeoGNN backbone network. Table 4 illustrates the influence of different objective combinations and the weight $w$ of law-of-cosines geometric consistency constraint. Although progressively incorporating semantic and geometric tasks generally enhances representational capability for macroscopic properties, as evidenced by the decrease in RMSE on FreeSolv from 1.868 to 1.821, unconstrained optimization can adversely affect performance on highly conformation-sensitive properties. Blindly fitting the coarse coordinates generated by empirical force fields such as MMFF94 forces the network to memorize spatial noise [56], which can lead to pronounced negative transfer. For instance, the RMSE on Lipophilicity dataset increases from 0.635 without pre-training to 0.641 under unconstrained pre-training.

However, introducing the law-of-cosines geometric consistency constraint, such as the Context + Graph + Geometry + Cosine configuration with $w = 4.5$, effectively resolves this dilemma and yields the best overall performance. By explicitly enforcing this geometric prior, the latent representations are compelled to align strictly with Euclidean geometric principles. This physics-guided constraint serves as a powerful structural filter, enabling our network to distill highly robust and invariant three-dimensional spatial representations from inherently noisy conformational inputs. Consequently, it successfully mitigates the spatial noise, reducing RMSE on Lipophilicity from the degraded 0.641 to an optimal 0.625, thereby surpassing the non-pre-trained baseline. Simultaneously, it optimizes the RMSE on FreeSolv from 1.821 down to its lowest1.634. Further ablation experiments on the weight of law-of-cosines loss function indicate that the choice of constraint strength $w$ has a substantial impact on final performance. An insufficient constraint with $w = 1$ fails to adequately filter the noise inherent in empirical force fields, modestly reducing the RMSE on Lipophilicity from 0.641 to 0.634. In contrast, an excessive constraint with $w = 9$ dominates the loss landscape. Such rigid over-regularization enforces strict mathematical alignment at the expense of capturing subtle chemical semantics, inducing a "gradient starvation" phenomenon for other objectives. As a result, while this extreme setting incidentally yields marginal gains on the relatively simple topological task ESOL by reducing the RMSE from 0.821 to 0.815, the errors on highly conformation-dependent tasks rebound. Specifically, the RMSE on Lipophilicity increases from 0.625 to 0.636, and the RMSE on FreeSolv increases from 1.634 to 1.680. Ultimately, setting $w = 4.5$ achieves a Pareto-optimal balance.

### 6.3 Ablation study of layer-SE and prediction head

To systematically disentangle the effects of feature modulation at different architectural stages, we conduct a comprehensive two-dimensional ablation study. By isolating the backbone layer-SE for intra-layer calibration within the encoder and the prediction head for decoding at the readout stage, we elucidate how various architectural combinations adapt to the specific inductive biases of heterogeneous chemical endpoints [57].

#### 6.3.1 Backbone layer-SE

To systematically evaluate the impact of different initialization strategies for the backbone layer-SE, the first three rows of Table 4 compare three distinct configurations, which are a fully unmodulated backbone (Disabled variant), a backbone with standard randomly initialized SE modules (Standard variant), and our proposed identity mapping initialization (Identity-Init strategy). This comparison clearly illustrates the severe "catastrophic forgetting" dilemma in fine-tuning and validates the effectiveness of our proposed initialization strategy [58].

By completely disabling the layer-SE, our network achieves the lowest errors on highly geometry-sensitive microscopic quantum tasks, such as QM7 and QM8, with MAEs of 60.35 and 0.0166, respectively. This superior performance stems from the fact that quantum mechanical properties (e.g., atomization energy) are strictly determined by precise Euclidean geometric structures. Having already internalized high-fidelity spatial representations during pre-training, our network benefits from directly passing these unaltered geometric embeddings to the decoder when downstream data are limited, thereby preserving spatial fidelity without introducing structural distortions. Because microscopic quantum properties are extremely sensitive to sub-Ångström geometric precision, the pre-trained spatial priors are fragile, and any modulation may introduce bias, making the unmodulated backbone the optimal choice among all pre-trained model variants for preserving structural integrity.

In contrast, inserting a randomly initialized SE module into the backbone network abruptly compresses and distorts the carefully pre-trained geometric channels during the initial fine-tuning phase. This negative transfer causes catastrophic performance degradation on highly sensitive quantum tasks, with the MAE on QM7 surging from 60.35 to 75.53. Although the aggressive random gating incidentally mitigates overfitting on small-scale FreeSolv dataset, yielding an RMSE of 1.502 likely through strong regularization on only 642 molecules, it severely undermines the structural stability required for general 3D modeling.

To elegantly address this dilemma, we propose an identity-Init strategy. By serving as a transparent identity mapping at initialization, it shields the pre-trained spatial knowledge from initial gradient shocks, thereby stabilizing the MAE on QM7 at 70.50. Furthermore, it enables the backbone network to gradually learn task-specific channel weights in subsequent training epochs, achieving the best performance on large-scale complex datasets such as QM9 with an MAE of 0.00683 and Lipophilicity with an RMSE of 0.621.

### 6.3.2 Contributions of SENet and FiLM prediction head

Under Identity-Init configuration for the backbone layer-SE, the last two rows of Table 5 examine the impact of dynamically decoding global graph representations on predictions at both macro- and micro-scales [59].

On ESOL benchmark, replacing the static MLP with SENet + MLP prediction head yields the best RMSE of 0.807. Aqueous solubility is primarily governed by the presence and distribution of specific polar pharmacophores (e.g., hydroxyl groups) rather than by precise sub-Ångström atomic coordinates. Here, SENet mechanism serves as an ideal “semantic filter” at the readout stage. It dynamically suppresses irrelevant geometric noise while selectively amplifying the latent channels corresponding to key hydrophilic functional groups.

The complete SenCos-GEM utilizes a dual-modulation prediction head (SENet + FiLM + MLP). By seamlessly integrating FiLM, which applies edge-conditioned affine transformations for feature scaling and shifting, with SENet’s channel-wise filtering, this prediction head acts as a versatile universal adapter. Critically, it avoids rigid overfitting to a single metric and instead dynamically adapts its inductive bias according to the global geometric context. For instance, on the highly sensitive QM7 dataset, the full model substantially enhances predictive performance via FiLM's geometric modulation, reducing MAE from 70.50 yielded by SENet + MLP head to 67.1. By forgoing marginal gains on isolated datasets, the complete dual-modulation architecture achieves a Pareto-optimal trade-off across the full spectrum of properties. This full configuration is selected as the baseline owing to its superior generalization and robustness on classification tasks, particularly attaining SOTA performance of 0.874 on PCBA benchmark, which is the most industrially relevant large-scale high-throughput screening dataset.

It is noteworthy that when the optimal feature-modulation configuration is selected independently for each regression task as Oracle configuration, in which QM7 task uses the from-scratch training result of 56.6 for SE-GeoGNN reported in Table 3, the overall relative improvement over GEM reaches 6.67%. Specifically, this Oracle configuration achieves relative error reductions of up to 19.98% on FreeSolv, 5.91% on Lipo, 8.45% on QM9, 3.90% on QM7, and 2.92% on QM8, with a minor trade-off of −1.13% on ESOL.

## 7. Conclusions

In this study, we propose SenCos-GEM, a novel SENet-calibrated and law-of-cosines-constrained geometry-enhanced molecular representation learning framework for accurate molecular property prediction. Extensive experiments on MoleculeNet benchmark datasets demonstrate that, compared with baseline methods such as GEM, SenCos-GEM substantially reduces predictive errors on regression tasks sensitive to three-dimensional conformations, such as FreeSolv, Lipophilicity, and QM9, while maintaining strong generalization on large-scale classification tasks such as PCBA. Comprehensive ablation studies validate the individual and synergistic contributions of cosine constraint, hierarchical SE modules, and prediction head. Additional diagnostic analyses further confirm our framework's enhanced sensitivity to stereoisomers and superior discriminative capability for conformationally perturbed structures, underscoring the effectiveness of proposed physical consistency constraints and adaptive modulation mechanisms.

The superior performance of SenCos-GEM stems from its core methodological innovations, which effectively address the key limitations of previous geometry-enhanced models. First, unlike existing approaches in which large-scale pre-training is vulnerable to geometric noise from empirical force fields, SenCos-GEM introduces a physics-constrained self-supervised pre-training strategy. By explicitly enforcing Euclidean relationships among bond lengths, bond angles, and interatomic distances through a geometric consistency loss based on the law of cosines, our framework extracts high-fidelity and mathematically invariant three-dimensional spatial priors from coarse 3D conformations without resorting to computationally prohibitive quantum chemical methods. Second, to mitigate catastrophic forgetting or negative transfer during fine-tuning on limited downstream data, our framework incorporates SE modules into each GeoGNN layer. These modules are intentionally omitted during pre-training and randomly initialized as identity mappings during fine-tuning, serving as lightweight, task-specific adapters that enable adaptive recalibration of channel-wise feature responses while preserving the rich pre-trained geometric priors. Finally, to overcome the rigidity of static MLP prediction heads, SenCos-GEM incorporates an advanced dual-modulation prediction head that integrates SENet-style channel gating mechanisms with FiLM conditioned on aggregated edge representations. This design enables dynamic and context-aware scaling and shifting of graph-level representations, thereby achieving superior task-specific mapping capabilities.

By leveraging physics-guided constraints and lightweight adaptive components, SenCos-GEM achieves explicit decoupling between general geometric pre-training and task-specific adaptation, paving the way for developing more reliable, generalizable, and robust molecular representation models. Future work will extend SenCos-GEM to more complex molecular tasks, such as protein-ligand binding affinity prediction, integrate it with de novo molecular generation models, and explore scaling to larger, chemically richer pre-training datasets as well as advanced architectures at billion-parameter scale, including Graph Transformers, Equivariant GNNs, and other molecular graph foundation models.

## Funding

This study is supported by Science and Technology Plan Project of Liaoning Province (Grant No. 2025-MSLH-351), Fundamental Research Funds for the Liaoning Universities (Grant No. LJ212410146026).

## Conflicts of interest

The authors declare that the research was conducted in the absence of any commercial or financial relationships that could be construed as a potential conflict of interest.

## Data availability

The source code and datasets are available online at

https://github.com/zhaoqi106/BTP-MFFGNN.

## Figure Legends

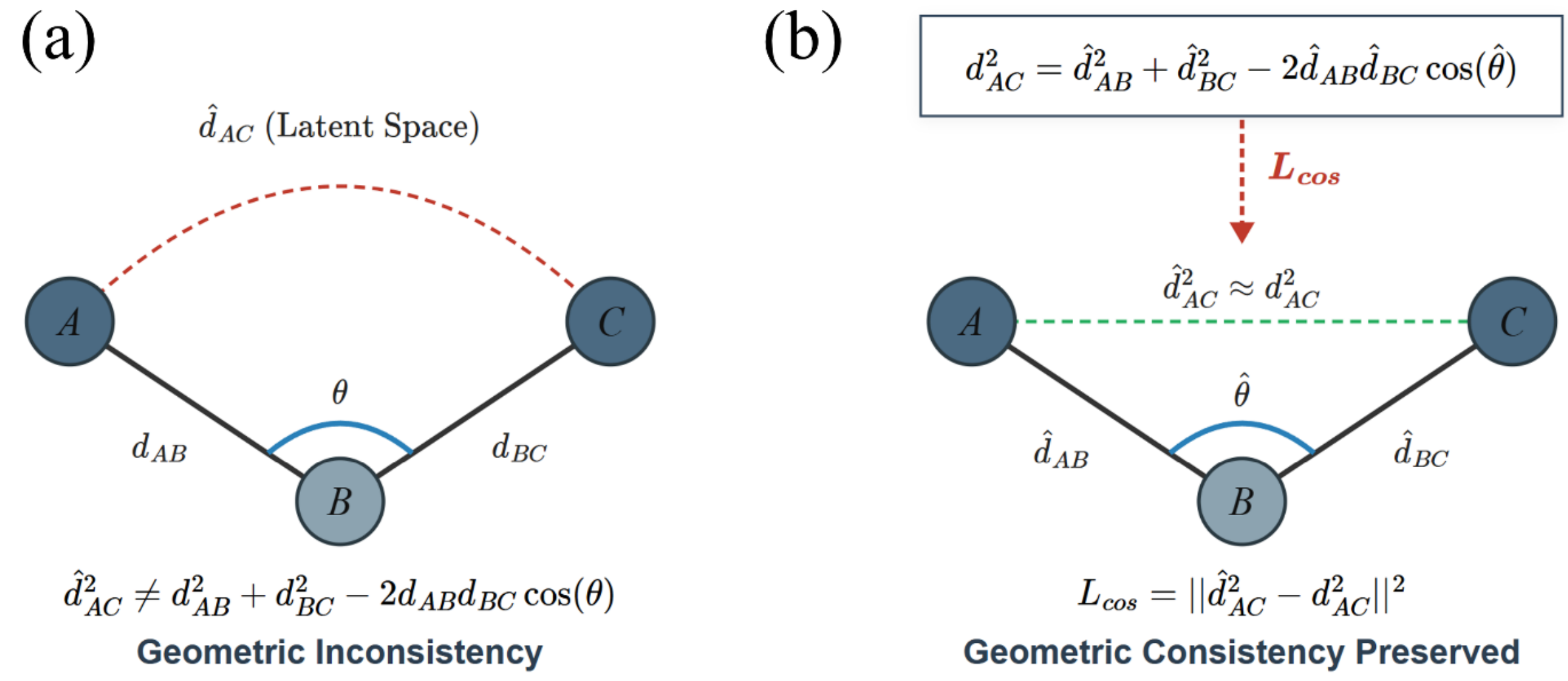

**Figure 1.** Comparison of geometric consistency in dual-graph models. (a) Traditional dual-graph models. Existing methods model local geometric features, such as bond distances and angles, independently. Consequently, the predicted latent distance between atoms *A* and *C* easily violates standard Euclidean geometry, leading to geometric inconsistency. (b) Our method constrained by the law of cosines. We explicitly integrate the law of cosines as a physical prior. Enforcing the geometric consistency loss $L_{cos}$ ensures that the latent representations rigorously respect standard Euclidean geometry.

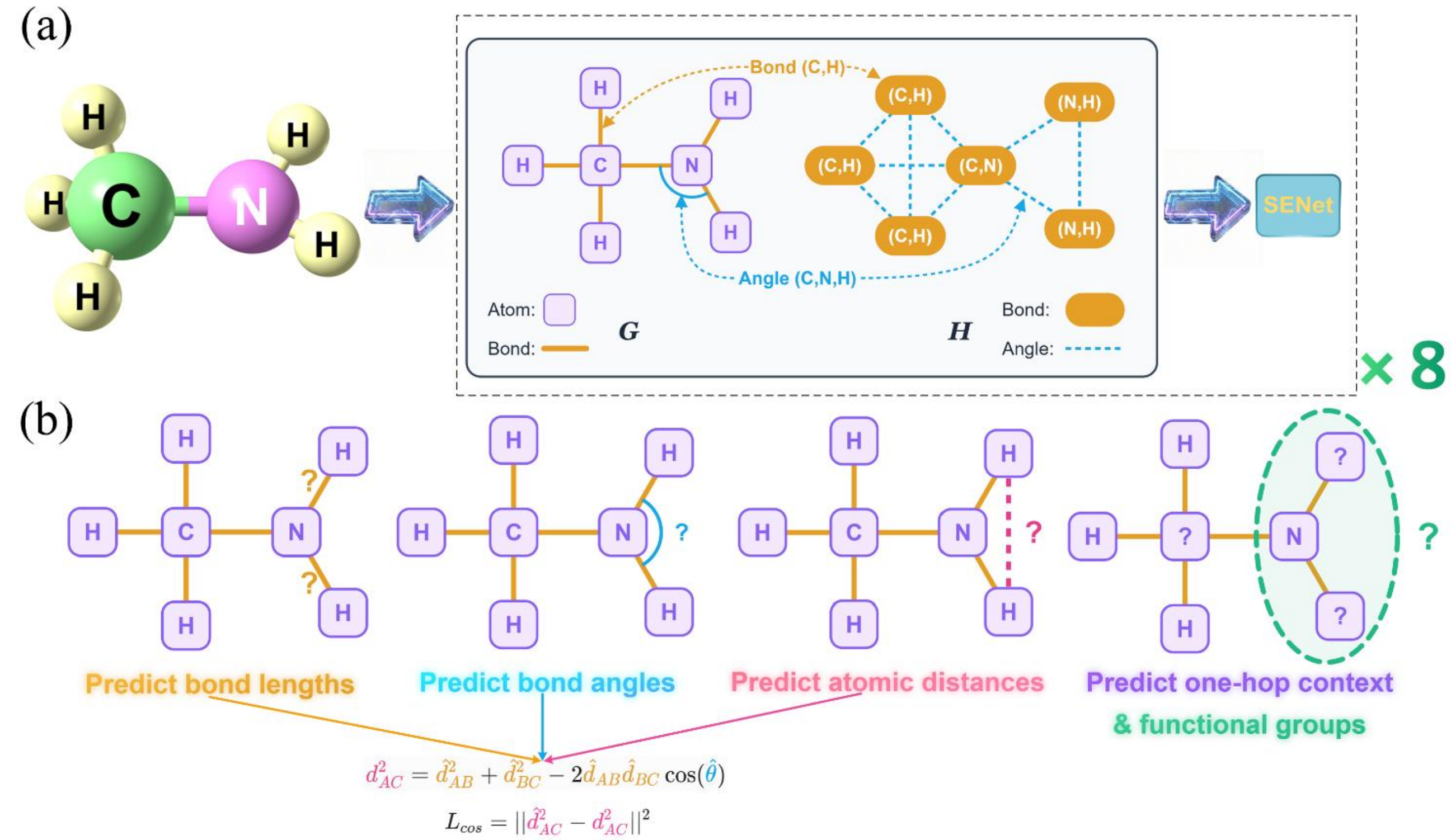

**Figure 2.** Overall framework and synergistic pre-training mechanism of SenCos-GEM. (a) The backbone of SE-GeoGNN. Through 8 stacked layers (×8), atom and bond representations are updated on graphs $G$ and $H$ respectively, with angle-aware bond features from $H$ passed to $G$ in the next layer. The double-dashed arcs explicitly indicate the mapping correspondence between elements across the dual graphs. Notably, SENet modules are bypassed during pre-training and identity-initialized for fine-tuning. (b) Physics-guided synergistic pre-training. Our model concurrently reconstructs local geometry, global spatial structure, and chemical context. Crucially, the geometric consistency loss $L_{\cos}$ organically integrates these isolated geometric tasks.

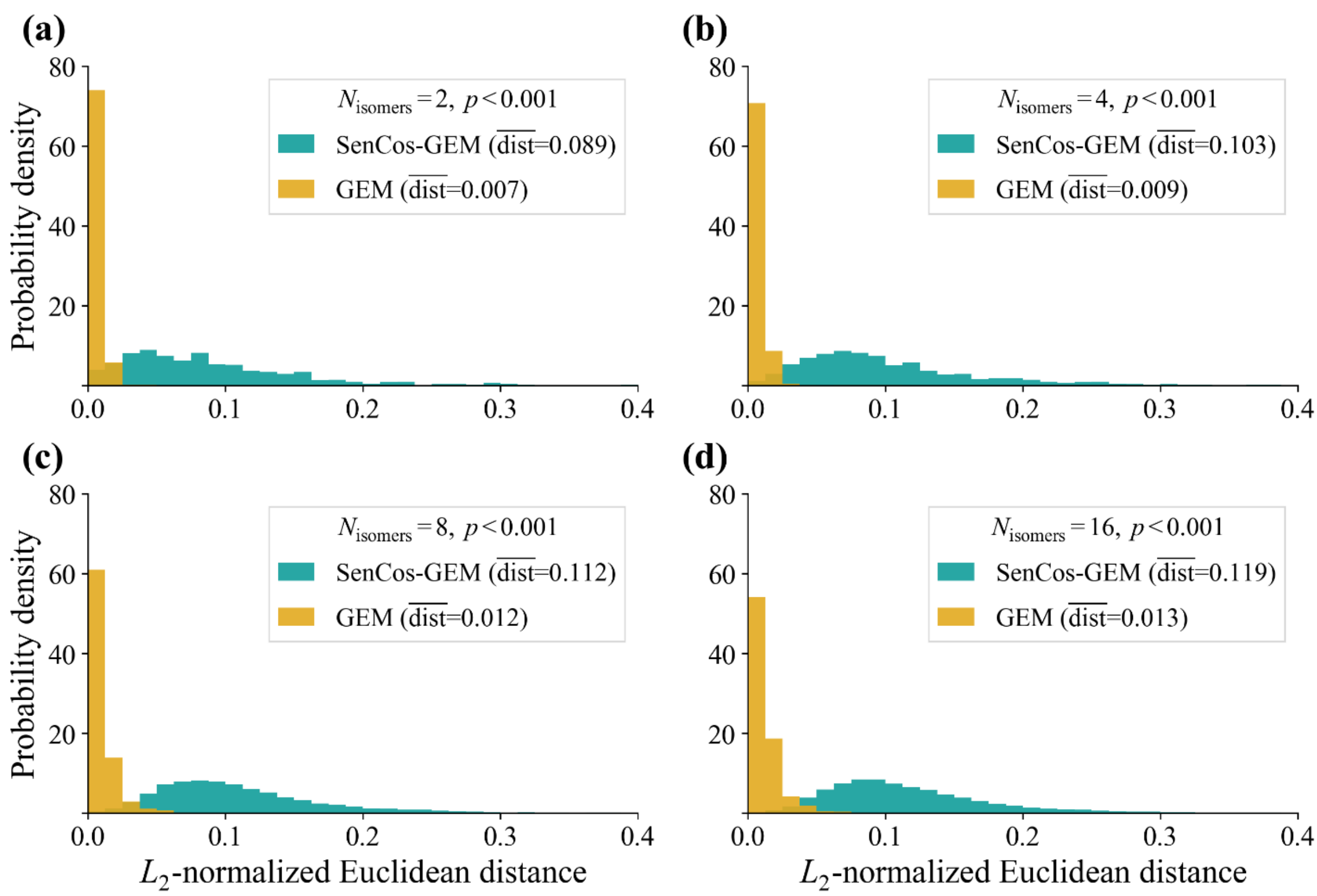

**Figure 3.** Ability of GEM and SenCos-GEM to distinguish stereoisomers. Pairwise Euclidean distance distributions are computed between $L_2$-normalized molecular representations of stereoisomers within the same molecular family.

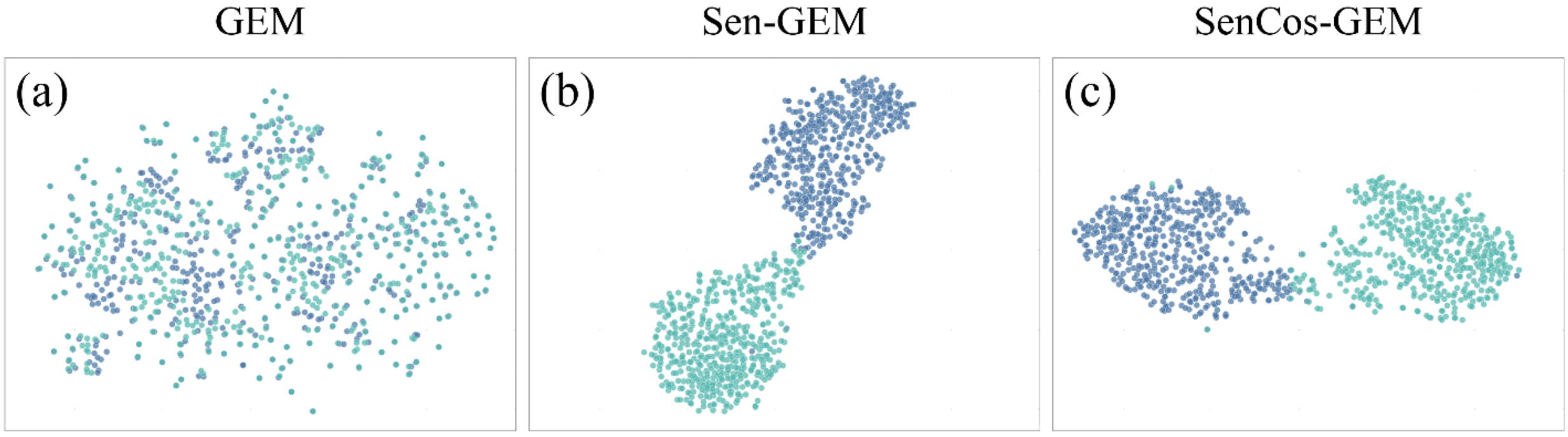

**Figure 4.** Representation-space separation of molecular conformational perturbations across GEM variants. For each molecule, embeddings are extracted from MMFF94-optimized conformer (blue points) and its noise-perturbed counterpart (green points), followed by t-SNE visualization. Compared with (a) GEM baseline yielding a DBI of 4.45, both (b) Sen-GEM and (c) SenCos-GEM show substantially clearer separation, reducing the DBI to 1.94 and 1.60, respectively.

## Table Legends

**Table 1.** Overall performance for regression tasks. The SOTA results are shown in bold.

| | RMSE↓ | | | MAE↓ | | |
|---|---|---|---|---|---|---|
| **Dataset** | **ESOL** | **FreeSolv** | **Lipo** | **QM7** | **QM8** | **QM9** |
| **No. molecules** | **1,128** | **642** | **4,200** | **6,830** | **21,786** | **133,885** |
| **No. prediction tasks** | **1** | **1** | **1** | **1** | **12** | **12** |
| D-MPNN | 1.050(0.008) | 2.082(0.082) | 0.683(0.016) | 103.5(8.6) | 0.0190(0.0001) | 0.00814(0.00001) |
| AttentiveFP | 0.877(0.029) | 2.073(0.183) | 0.721(0.001) | 72.0(2.7) | 0.0179(0.0001) | 0.00812(0.00001) |
| N-Gram$_{RF}$ | 1.074(0.107) | 2.688(0.085) | 0.812(0.028) | 92.8(4.0) | 0.0236(0.0006) | 0.01037(0.00016) |
| N-Gram$_{XGB}$ | 1.083(0.082) | 5.061(0.744) | 2.072(0.030) | 81.9(1.9) | 0.0215(0.0005) | 0.00964(0.00031) |
| PretrainGNN | 1.100(0.006) | 2.764(0.002) | 0.739(0.003) | 113.2(0.6) | 0.0200(0.0001) | 0.00922(0.00004) |
| GROVER$_{base}$ | 0.983(0.090) | 2.176(0.052) | 0.817(0.008) | 94.5(3.8) | 0.0218(0.0004) | 0.00984(0.00055) |
| GROVER$_{large}$ | 0.895(0.017) | 2.272(0.051) | 0.823(0.010) | 92.0(0.9) | 0.0224(0.0003) | 0.00986(0.00025) |
| GEM | [a]**0.798(0.029)** | [a]1.877(0.094) | [a]0.660(0.008) | [a]**58.9(0.8)** | [a]0.0171(0.0001) | [a]0.00746(0.00001) |
| SenCos-GEM | 0.821(0.012) | **1.634(0.098)** | **0.625(0.008)** | 67.1(5.1) | **0.0168(0.0001)** | **0.00685(0.00008)** |

[a]These cells indicate the previous SOTA results.

**Table 2.** Overall performance for classification tasks. The SOTA results are shown in bold.

| **Dataset** | **BACE** | **BBBP** | **ClinTox** | **SIDER** | **Tox21** | **ToxCast** | **HIV** | **MUV** | **PCBA** | **Avg** |
|---|---|---|---|---|---|---|---|---|---|---|
| **No. molecules** | **1,513** | **2,039** | **1,478** | **1,427** | **7,831** | **8,575** | **41,127** | **93,087** | **437,929** | |
| **No. prediction tasks** | **1** | **1** | **2** | **27** | **12** | **617** | **1** | **17** | **128** | |
| D-MPNN | 0.809(0.006) | 0.710(0.003) | [a]**0.906(0.006)** | 0.570(0.007) | 0.759(0.007) | 0.655(0.003) | 0.771(0.005) | 0.786(0.014) | 0.862(0.001) | 0.759 |
| AttentiveFP | 0.784(0.022) | 0.643(0.018) | 0.847(0.003) | 0.606(0.032) | 0.761(0.005) | 0.637(0.002) | 0.757(0.014) | 0.766(0.015) | 0.801(0.014) | 0.734 |
| N-Gram$_{RF}$ | 0.779(0.015) | 0.697(0.006) | 0.775(0.040) | 0.668(0.007) | 0.743(0.004) | [b]— | 0.772(0.001) | 0.769(0.007) | [b]— | — |
| N-Gram$_{XGB}$ | 0.791(0.013) | 0.691(0.008) | 0.875(0.027) | 0.655(0.007) | 0.758(0.009) | [b]— | 0.787(0.004) | 0.748(0.002) | [b]— | — |
| PretrainGNN | 0.845(0.007) | 0.687(0.013) | 0.726(0.015) | 0.627(0.008) | 0.781(0.006) | 0.657(0.006) | 0.799(0.007) | 0.813(0.021) | 0.860(0.001) | 0.755 |
| GROVER$_{base}$ | 0.826(0.007) | 0.700(0.001) | 0.812(0.030) | 0.648(0.006) | 0.743(0.001) | 0.654(0.004) | 0.625(0.009) | 0.673(0.018) | 0.765(0.021) | 0.716 |
| GROVER$_{large}$ | 0.810(0.014) | 0.695(0.001) | 0.762(0.037) | 0.654(0.001) | 0.735(0.001) | 0.653(0.005) | 0.682(0.011) | 0.673(0.018) | 0.830(0.004) | 0.722 |
| GEM | [a]0.856(0.011) | [a]0.724(0.004) | 0.901(0.013) | [a]**0.672(0.004)** | [a]0.781(0.001) | [a]**0.692(0.004)** | [a]**0.806(0.009)** | [a]**0.817(0.005)** | [a]0.866(0.001) | [a]**0.791** |
| SenCos-GEM | **0.861(0.013)** | **0.725(0.005)** | 0.894(0.021) | 0.650(0.003) | **0.783(0.006)** | 0.681(0.006) | 0.777(0.007) | 0.797(0.023) | **0.874(0.001)** | 0.782 |

[a]These cells indicate the previous SOTA results. [b]For N-Gram on ToxCast and PCBA, this evaluation is omitted due to computationally prohibitive training costs.

**Table 3.** Performance of different GNN architectures for regression tasks. The SOTA results are shown in bold.

| | | RMSE↓ | | | MAE↓ | |
|---|---|---|---|---|---|---|
| **Method** | **ESOL** | **FreeSolv** | **Lipo** | **QM7** | **QM8** | **QM9** |
| GIN | 1.067(0.051) | 2.346(0.122) | 0.757(0.022) | 110.3(7.2) | 0.0199(0.0002) | 0.00886(0.00005) |
| GAT | 1.556(0.085) | 3.559(0.050) | 1.021(0.029) | 103.0(4.4) | 0.0224(0.0005) | 0.01117(0.00018) |
| GCN | 1.211(0.052) | 3.174(0.308) | 0.773(0.007) | 100.0(3.8) | 0.0203(0.0005) | 0.00923(0.00019) |
| D-MPNN | 1.050(0.008) | 2.082(0.082) | 0.683(0.016) | 103.5(8.6) | 0.0190(0.0001) | 0.00814(0.00009) |
| AttentiveFP | 0.877(0.029) | 2.073(0.183) | 0.721(0.001) | 72.0(2.7) | 0.0179(0.0001) | 0.00812(0.00001) |
| GTransformer | 2.298(0.118) | 4.480(0.155) | 1.112(0.029) | 161.3(7.1) | 0.0361(0.0008) | 0.00923(0.00019) |
| SGCN | 1.629(0.001) | 2.363(0.050) | 1.021(0.013) | 131.3(11.6) | 0.0285(0.0005) | 0.01459(0.00055) |
| DimeNet | 0.878(0.023) | 2.094(0.118) | 0.727(0.019) | 95.6(4.1) | 0.0215(0.0003) | 0.01031(0.00076) |
| HMGNN | 1.390(0.073) | 2.123(0.179) | 2.116(0.473) | 101.6(3.2) | 0.0249(0.0004) | 0.01239(0.0001) |
| GeoGNN | [a]**0.832(0.010)** | [a]1.857(0.071) | [a]0.666(0.015) | [a]59.0(3.4) | [a]0.0173(0.0004) | [a]0.00746(0.00003) |
| SE-GeoGNN | 0.839(0.024) | **1.765(0.057)** | **0.658(0.010)** | **56.6(0.6)** | **0.0168(0.0002)** | **0.00695(0.00008)** |

[a]The cells indicate the previous SOTA results.

**Table 4.** Performance of GeoGNN with different pre-training strategies for regression tasks. The SOTA results are shown in bold.

| | RMSE↓ | | | MAE↓ | | |
|---|---|---|---|---|---|---|
| **Pre-train Method** | **ESOL** | **FreeSolv** | **Lipo** | **QM7** | **QM8** | **QM9** |
| Without pre-train | $0.855_{(0.027)}$ | $1.868_{(0.042)}$ | $0.635_{(0.007)}$ | $\mathbf{59.29_{(1.9)}}$ | $\mathbf{0.0166_{(0.0001)}}$ | $0.00688_{(0.00008)}$ |
| Graph + Geometry + Cosine (*w* = 2) | $0.844_{(0.013)}$ | $1.888_{(0.131)}$ | $0.640_{(0.007)}$ | $74.60_{(3.7)}$ | $\mathbf{0.0166_{(0.0002)}}$ | $0.00693_{(0.00002)}$ |
| Geometry + Cosine (*w* = 4.5) | $0.854_{(0.031)}$ | $1.949_{(0.137)}$ | $0.637_{(0.017)}$ | $68.11_{(3.1)}$ | $0.0170_{(0.0004)}$ | $0.00688_{(0.00008)}$ |
| Context + Graph + Geometry | $0.835_{(0.021)}$ | $1.821_{(0.044)}$ | $0.641_{(0.017)}$ | $74.78_{(5.3)}$ | $0.0168_{(0.0002)}$ | $0.00695_{(0.00004)}$ |
| Context + Graph + Geometry + Cosine (*w* = 1) | $0.821_{(0.006)}$ | $1.711_{(0.104)}$ | $0.634_{(0.002)}$ | $72.61_{(6.5)}$ | $0.0169_{(0.0003)}$ | $0.00694_{(0.00007)}$ |
| Context + Graph + Geometry + Cosine (*w* = 9) | $\mathbf{0.815_{(0.015)}}$ | $1.680_{(0.075)}$ | $0.636_{(0.013)}$ | $74.00_{(1.6)}$ | $0.0168_{(0.0004)}$ | $0.00696_{(0.00003)}$ |
| Context + Graph + Geometry + Cosine (*w* = 4.5) | $0.821_{(0.012)}$ | $\mathbf{1.634_{(0.098)}}$ | $\mathbf{0.625_{(0.008)}}$ | $67.1_{(5.1)}$ | $0.0168_{(0.0001)}$ | $\mathbf{0.00685_{(0.00008)}}$ |

**Table 5.** Ablation of feature modulation for regression tasks. The SOTA results are shown in bold.

| | | RMSE↓ | | | MAE↓ | | |
|---|---|---|---|---|---|---|---|
| **Backbone layer-SE** | **Prediction Head** | **ESOL** | **FreeSolv** | **Lipo** | **QM7** | **QM8** | **QM9** |
| Disabled | MLP | $0.823_{(0.032)}$ | $1.558_{(0.063)}$ | $0.623_{(0.006)}$ | $\mathbf{60.35_{(2.3)}}$ | $\mathbf{0.0166_{(0.0001)}}$ | $0.00690_{(0.00005)}$ |
| Standard | MLP | $0.833_{(0.021)}$ | $\mathbf{1.502_{(0.039)}}$ | $0.635_{(0.007)}$ | $75.53_{(3.4)}$ | $\mathbf{0.0166_{(0.0002)}}$ | $0.00688_{(0.00008)}$ |
| Identity-Init | MLP | $0.827_{(0.016)}$ | $1.593_{(0.056)}$ | $\mathbf{0.621_{(0.008)}}$ | $70.22_{(4.6)}$ | $0.0167_{(0.0001)}$ | $\mathbf{0.00683_{(0.00002)}}$ |
| Identity-Init | SENet + MLP | $\mathbf{0.807_{(0.021)}}$ | $1.583_{(0.056)}$ | $0.639_{(0.013)}$ | $70.50_{(5.8)}$ | $0.0168_{(0.0001)}$ | $0.00691_{(0.00004)}$ |
| Identity-Init | SENet + FiLM + MLP | $0.821_{(0.012)}$ | $1.634_{(0.098)}$ | $0.625_{(0.008)}$ | $67.1_{(5.1)}$ | $0.0168_{(0.0001)}$ | $0.00685_{(0.00008)}$ |